\documentclass{article}
\pdfoutput=1
\usepackage[final, nonatbib]{neurips_2021}

\usepackage[utf8]{inputenc} % allow utf-8 input
\usepackage[T1]{fontenc}    % use 8-bit T1 fonts
\usepackage[hidelinks]{hyperref}
\usepackage{url}            % simple URL typesetting
\usepackage{booktabs}       % professional-quality tables
\usepackage{amsfonts}       % blackboard math symbols
\usepackage{nicefrac}       % compact symbols for 1/2, etc.
\usepackage{microtype}      % microtypography
\usepackage{xcolor}         % colors
\usepackage{commath}
\usepackage{graphicx}
\usepackage{wrapfig}
\usepackage[font=small,labelfont=bf]{caption}
\usepackage{color, colortbl}
\definecolor{Gray}{gray}{0.9}

\title{Domain-Agnostic Clustering with Self-Distillation}
\author{%
  Mohammed Adnan$^{1,2}$, Yani A.\ Ioannou$^{3}$, Chuan-Yung Tsai$^{2}$, Graham~W.~Taylor$^{1,2}$\\
  
  \And 
  \\
  
  $^1$University of Guelph, $^2$Vector Institute for AI, $^3$University of Calgary\\
  \vspace{-100pt}
  \And 
  \\
  \texttt{\{}\href{mailto:madnan01@uoguelph.ca}{\texttt{madnan01}}, \href{mailto:gwtaylor@uoguelph.ca}{\texttt{gwtaylor}}\texttt{\}}\texttt{@uoguelph.ca}, \\ \href{mailto:yani.ioannou@ucalgary.ca}{\texttt{yani.ioannou@ucalgary.ca}}, \href{mailto:kenyon.tsai@vectorinstitute.ai}{\texttt{kenyon.tsai@vectorinstitute.ai}}\\
  
}

\begin{document}
\maketitle

\begin{abstract}
Recent advancements in self-supervised learning have reduced the gap between supervised and unsupervised representation learning. However, most self-supervised and deep clustering techniques rely heavily on data augmentation, rendering them ineffective for many learning tasks where insufficient domain knowledge exists for performing augmentation. We propose a new self-distillation based algorithm for domain-agnostic clustering. 
Our method builds upon the existing deep clustering frameworks and requires no separate student model. The proposed method outperforms existing domain agnostic (augmentation-free) algorithms on CIFAR-10. We empirically demonstrate that knowledge distillation can improve unsupervised representation learning by extracting richer `dark knowledge' from the model than using predicted labels alone. 
Preliminary experiments also suggest that self-distillation improves the convergence of DeepCluster-v2.
\end{abstract}

\section{Introduction}
In recent years, the representation learning community has put much focus on learning with minimal labeled data or even no labeled data. Self-Supervised Learning (SSL) refers to the class of learning algorithms that use no human labels. Rather, SSL  obtains self-supervisory signals from the data itself, often leveraging its underlying structure. These self-supervisory signals are obtained using pretext tasks such as predicting the rotation of images~\cite{gidaris2018unsupervised}, solving jigsaw puzzles~\cite{noroozi2017unsupervised},  mixup~\cite{zhang2018mixup}, and colorization~\cite{zhang2016colorful}. Most of the methods use contrastive energy-based pretext tasks. SimCLR~\cite{chen2020simple} proposed a simple framework for learning visual representation by minimizing the distance between different views (augmentation) of the same instance via contrastive loss. Since computing embeddings for the entire dataset is not feasible, memory bank based approaches have also been proposed. He~et~al.~\cite{he2020momentum} proposed MoCo, a dynamic dictionary-based framework for unsupervised learning of visual representations.
Other methods such as DeepCluster~\cite{caron_deep_2019}, ClusterFit~\cite{yan2019clusterfit}, SwAV~\cite{caron_unsupervised_2021}, Barlow Twins~\cite{zbontar2021barlow}, and BYOL~\cite{grill2020bootstrap} are non-contrastive and use feature clustering to learn representations. They use various pretext tasks such as computing target embeddings to group similar images or minimize the redundancy between the individual components of the embedding vector.
However, designing pretext tasks requires domain knowledge and hinders the application of SSL as an out-of-the-box solution for arbitrary domains where domain knowledge is not available. 
Thus, it is necessary to develop domain agnostic (data augmentation free) clustering and SSL algorithms. 

In this work, we propose a novel self-distillation based domain-agnostic clustering algorithm which does not require any domain knowledge for designing pretext tasks or data augmentation. %We experimented on CIFAR-10~\cite{Krizhevsky09learningmultiple} and achieved state-of-the-art performance compared to existing non-contrastive clustering methods.%
We also demonstrate that knowledge distillation can provide further self-supervisory signals from soft labels (i.e.~dark knowledge).

\section{Background}

\noindent\textbf{Knowledge Distillation. }
Knowledge Distillation (KD) is a model compression method in which a smaller `student' model is trained to mimic the behavior of a large `teacher' model by minimizing the loss on the output class probabilities (soft labels) of the large model~\cite{hinton2015distilling}. %Hinton~et~al. proposed to use \emph{softmax temperature ($T$)} to smooth out class probabilities.% 
It has been found that the smaller model achieves similar, or often better performance than the original model. This~behavior has been attributed to the presence of `dark knowledge' present in the soft labels. Soft labels provide much more information about the semantic information present in the image. For example, given a dog image from CIFAR-10~\cite{Krizhevsky09learningmultiple}, the class probability of the image being a cat will be much higher than the class probability of the image being a car. Thus, the softmax values give additional hints to the network that images of dogs and cats contain similar semantic information. Knowledge distillation also improves the loss landscape and helps find flat minima, which in turn improves generalization. Distillation has been shown to amplify regularization in the Hilbert space, and thus, it improves generalization~\cite{mobahi2020selfdistillation}.
In \emph{self-distillation} rather than using a separate student and teacher network, a single model is used to extract `dark knowledge'~\cite{zhang2019teacher,hou2019learning,zhang2020selfdistillation,yang2018snapshot, Phuong2019DistillationBasedTF,lan2018selfreferenced}. The main objective in self distillation is to distill the knowledge from the deeper layers to the shallow layers of the network~\cite{Gou_2021}.

\noindent\textbf{DeepCluster. }
Caron~et~al.\ proposed Deep Clustering~\cite{caron_deep_2019} (DeepCluster) to jointly learn the neural network and cluster the resulting features by iteratively applying k-means. DeepCluster can learn generalizable features in an unsupervised manner using clustering as a pretext task. Since convolutional layers encode a strong prior for learning natural images, a randomly initialized CNN achieves 12\% accuracy on ImageNet while a random prediction would be 0.1\% for a randomly-initialized fully connected model~\cite{noroozi2017unsupervised}.
DeepCluster exploits this weak signal to bootstrap the discriminative power of a CNN. Given a CNN mapping denoted by $f_\theta$, a classifier parameterized by $g_{w}$ and a training set of $X = \{x_1, x_2\dots, x_n$\}, it clusters the output of a CNN and subsequently clusters the output features to optimize: 

\begin{equation}
    \min_{\theta, W} \frac{1}{N} \sum_{n=1}^{N} \ell(g_{w}(f_{\theta}(x_n), y_n ).
    \label{eq:deep_clus}
\end{equation}

This process is repeated iteratively until the model converges. After training, the classifier $g_w$ is discarded, and the CNN $f_{\theta}$ can be used for downstream learning tasks. Since there is no mapping between assignments in two consecutive epochs, the classifier ($g_w$) is re-initialized after each epoch.
However, training DeepCluster is not trivial, as with any method that attempts to jointly learn a discriminative classifier and labels. Pseudo-labels can collapse to one class, or it is possible that some of the clusters will be empty. The collapse happens when the CNN outputs similar features for different inputs. In either case, the training loss decreases without progress in learning discriminative features. DeepCluster-v2~\cite{caron_unsupervised_2021} uses centroids given by spherical k-means to initialize the classifier weights after each epoch which improves overall stability. %In our experiments we build upon and compare with DeepCluster-v2, and show that our method significantly improves training stability.%

\noindent\textbf{Domain agnostic clustering. }
Current state-of-the-art Self-Supervised Learning (SSL) and clustering algorithms use data augmentation either for learning contrastive representations or as a regularizer. However, in many scenarios, data augmentation techniques cannot be used. For example, \emph{color jittering} commonly used in SSL cannot be used on black and white x-ray images~\cite{tamkin2021dabs}, and \emph{random cropping}~\cite{Takahashi_2020} is not relevant to histopathology images where the region of interest is significantly smaller than the image itself. Thus, it is important to develop domain agnostic SSL and clustering algorithms that don't require the design of domain-aware components such as augmentations.  %\textcolor{red}{add background for existing domain agnostic methods}
Recently, some methods have been proposed for domain agnostic SSL. Verma~et~al.~\cite{verma_towards_2021} proposed a domain agnostic algorithm by using mixup noise at the hidden-state level. Viewmaker Network~\cite{tamkin_viewmaker_2021} uses a generative model to generate different views or augmentation, which can then be used for domain agnostic SSL.

\noindent\textbf{Impact of data augmentation on clustering. }
For SSL techniques where data augmentation is not a critical part of the supervisory signal, one way to achieve domain agnosticism is to remove the augmentation, incurring a decrease in accuracy. However, SSL methods rely heavily on data augmentation, and their performance decreases significantly after removing data augmentation. Various studies have shown the impact of data augmentation on SSL and clustering techniques. Deshmukh~et~al. in~\cite{deshmukh_representation_2021} noted that by just removing \emph{random crop} from the data augmentation pipeline, the accuracy on CIFAR-10~\cite{Krizhevsky09learningmultiple} drops from $84.59\%$ to $29.88\%$. 
Tao~et~al.~\cite{tao_clustering-friendly_2021} also reported a decrease in the accuracy from $81.5\%$ to $23.6\%$ after removing data augmentations.  Chen~et~al. also observed that data augmentation is crucial for SimCLR to learn good visual representations~\cite{chen2020simple}.
One explanation of the decrease in the accuracy after removing data augmentations could be attributed to the generalization property of data augmentations. Similar to explicit regularization techniques such as weight decay and dropout, data augmentations have been shown to act as an implicit regularizer that improves generalization~\cite{hernandezgarcia2020data}.

\begin{figure}[t]
    \centering
    \includegraphics[width=0.80\linewidth]{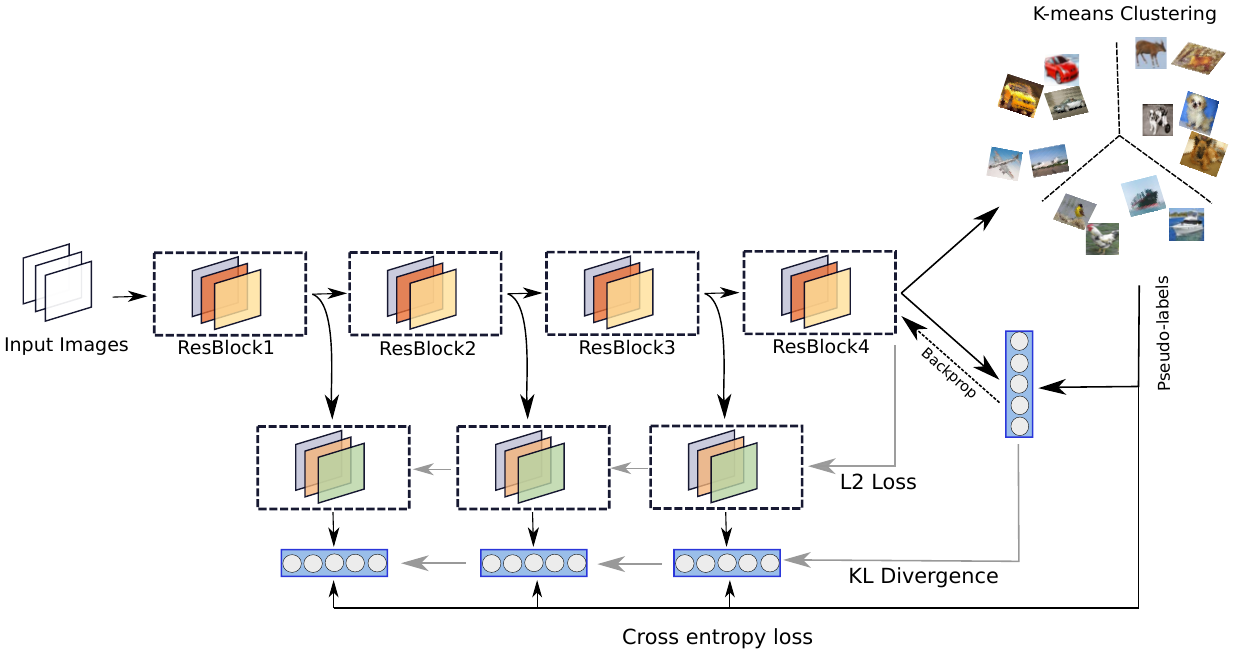}
    \caption{Schematic overview of our method. Bottleneck branches between ResBlocks act as student models and the deepest layer acts as a teacher model. Arrows in grey denote distillation loss (KL divergence and L2 loss). Each classifier has cross-entropy loss from the pseudo-labels given by k-means.}
    \label{fig:method}
\end{figure}

\section{Methodology}

\noindent\textbf{Motivation. } The underlying principle behind our proposed method is that semantically similar images of different classes are often embedded close in Euclidean space and are consequently misclassified by k-means. However, soft labels may provide more information about semantically similar classes, and thus by using self-distillation, the model can be provided with further information to distinguish those inputs. Moreover, self-distillation can partially substitute the regularizing effect of data augmentation and help find flat minima, which in turn improves generalization~\cite{mobahi2020selfdistillation}.

\noindent\textbf{Problem Statement. } Given an unlabelled dataset $X=\{x_1, x_2,\dots x_n\}$, our objective is to learn generalizable features $f_{\theta}(x_i)$ without using any domain-specific data augmentation, which can then be used for learning downstream tasks by only training a linear classifier $g_{w}$. 

\noindent\textbf{Proposed Approach. }
Our proposed method uses the DeepCluster-v2~\cite{caron_unsupervised_2021} framework while also introducing a self-distillation loss. We use ResNet~\cite{he2015deep} as the CNN for extracting features and introduce a self-distillation loss~\cite{zhang2019teacher}. The ResNet architecture is modified to have an additional three bottleneck branches. Secondly, an auxiliary classifier is added on top of bottleneck branches as shown in the~\autoref{fig:method}. During the training phase, all three bottleneck classifiers ($q_{i}, i=1,\ldots,3)$ along with the original classifier ($q_{c}$) are utilized. Bottleneck classifiers are trained as student models via distillation from the deepest classifier, which acts as the teacher model. To improve the overall performance, Zhang~et~al.~\cite{zhang2019teacher} introduced three losses:

\begin{enumerate}

    \item Loss Source 1:
    All bottleneck classifiers (student models) have cross entropy loss from the pseudo-labels obtained from k-means clustering denoted by $y_{k}(x)$: 
    \begin{equation}
        \mathcal{L}_{i} =  \sum_{\forall x} y_{k}(x) \log(q_{i}(x)) \hspace{0.7cm} ;i\in \{1,2,3,c\}. %CrossEntropy(q_{i},y_{kmeans})
    \end{equation}
    % \begin{equation}
    %     \mathcal{L}_{DC} = \sum_{\forall x} y_{k}(x)\log(q_{c}(x))
    % \end{equation}

    \item Loss Source 2: Kullback-Leibler (KL) divergence loss from the softmax output of the deepest classifier (teacher model). In this way, hidden knowledge from the softmax is infused in to the hidden layers for learning better representations:
    \begin{equation}
        \mathcal{L}_{KL} = \sum_{\forall x}q_{c}(x)\log(\frac{q_{c}(x)}{q_i(x)}). 
    \end{equation}
    
    \item Loss Source 3: L2 loss between the bottleneck feature map and the deepest layer features is added to provide implicit knowledge or hints to the bottleneck classifiers:
    \begin{equation}
       \mathcal{L}_{hints} = \norm{F_i - F_C}_2.
    \end{equation}
    
\end{enumerate}
During training, all three distillation losses are combined with the DeepCluster loss ($L_{Dc}$). The overall loss is given as:
\begin{equation}
    \mathcal{L}_{total} = \mathcal{L}_{c} + (1-\alpha). \sum_{i=1,2,3}\mathcal{L}_{i} + \alpha . \mathcal{L}_{KL} + \lambda . \mathcal{L}_{hints}  .
    \label{eq:distill_total}
\end{equation}

% At inference time, all the bottleneck branches are discarded and only the deepest layer is used to extract features. 
% We briefly discuss limitations of our method in~\autoref{sec:conc}.

\section{Experiments}

We evaluate the performance of our approach on the CIFAR-10 dataset~\cite{Krizhevsky09learningmultiple}. CIFAR-10 contains 50,000 labeled training images and 10,000 test images belonging to ten different classes. We train our model on 50,000 training images without using any ground labels. In contrast to most existing unsupervised learning algorithms, we do not use any domain knowledge for data augmentations. The trained model can be used to extract general-purpose features for downstream machine learning tasks. We evaluate the trained model on 10,000 images. Implementation details are provided in the~\autoref{sec:appendix}.

\noindent\textbf{Evaluation Metric. }
To evaluate the trained model ($f_{\theta}$), we freeze the network and train a linear classifier on top of the frozen CNN using training samples from the CIFAR-10. Test data is used to evaluate the accuracy, which quantitatively measures the quality of the features learned in an unsupervised manner within the frozen CNN layers.

\noindent\textbf{Results. }
We compare our proposed approach with existing domain agnostic or data augmentation-free clustering methods. Since most of the current clustering methods use data augmentation either for contrastive learning or as a regularizer, we compare with the performance of existing methods when trained without data augmentation.
Our method outperforms other methods on CIFAR-10 and shows significant improvement over DeepCluster-v2 as shown in~\autoref{tab:result}.

\begin{minipage}[b]{0.49\textwidth}
    \centering

  \begin{tabular}[b]{ll}
  
    \toprule

    \textbf{Method  }  & \textbf{Accuracy} \\
    \midrule
    
    ID~\cite{tao_clustering-friendly_2021}                              &18.7\%  \\
    IDFD~\cite{tao_clustering-friendly_2021}                            & 23.6\% \\
    ConCURL~\cite{deshmukh_representation_2021}                         & 29.88\% \\
    DeepCluster-v2~\cite{caron_unsupervised_2021}                    & $33.27 \pm 0.06$ \% \\
    %\rowcolor{Gray}
    DeepCluster+KD (ours)         & $\mathbf{38.00\pm 0.34\%}$  \\       
    \bottomrule

  \end{tabular}
\captionof{table}{Domain Agnostic Clustering on CIFAR-10. DeepCluster was trained with no data augmentation.}
      \label{tab:result}

\end{minipage}
\hfill
\begin{minipage}[b]{0.49\textwidth}
    \centering
\includegraphics[width=0.7\textwidth]{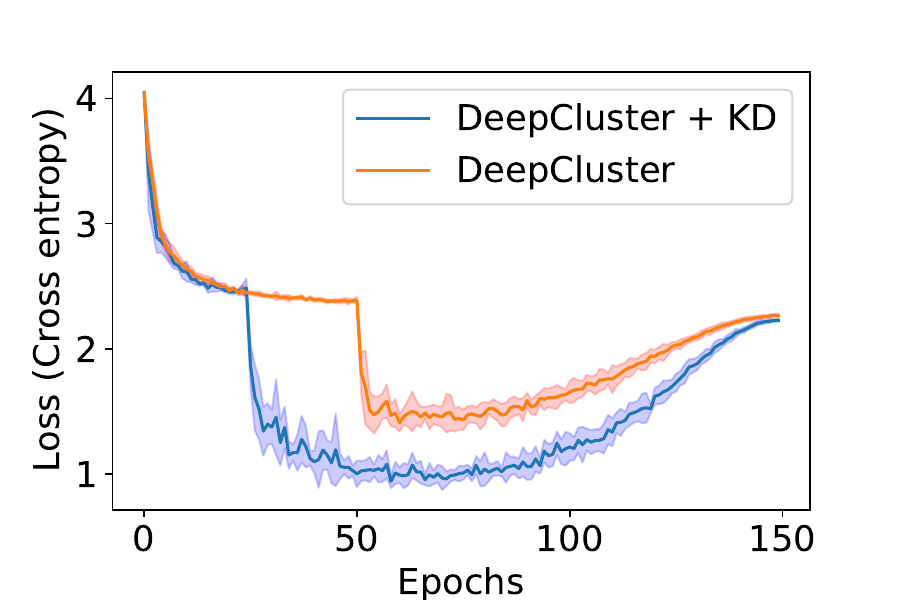}
\captionof{figure}{Evolution of the training loss. Proposed method converges faster than the DeepCluster-v2.}
\label{fig:epochs}
\end{minipage}

\noindent\textbf{Stability. } Preliminary experiments suggest that knowledge distillation helps in faster convergence and stability of DeepCluster training. Evaluation of the training loss with respect to the number of epochs is shown in~\autoref{fig:epochs}. All the training runs with different random initializations followed a similar trend suggesting improved stability and convergence due to self-distillation.

% \begin{figure}
%     \centering
%     \includegraphics[width=0.6\linewidth]{epochs.eps}
%     \caption{Evolution of the training loss. Proposed method converges faster than the DeepCluster-v2.}
%     \label{fig:epochs}
% \end{figure}

\section{Conclusions and future work}
In this work, we proposed that extracting hidden dark knowledge via self-distillation from the dataset further improves the performance of unsupervised clustering algorithms. Preliminary experiments on CIFAR-10 show that self-distillation improves the existing state-of-the-art method by $4.5\%$. 
In subsequent work, we would also benchmark the proposed algorithm on ImageNet and on the DABS: Domain Agnostic Benchmark for Self-Supervised Learning dataset~\cite{tamkin2021dabs} to study the benefits of distillation on large datasets. %It would be interesting to extend our work to contrastive clustering learning based SSL techniques.%

\label{sec:conc}
%%%%%%%%%%%%%%%%%%%%%%%%%%%%%%%%%%%%%%%%%%%%%%%%%%%%%%%%%%%%
\newpage
\bibliography{ref_ssl,ref_general}
\bibliographystyle{ieeetr}

\newpage
\appendix

\section{Appendix}
\label{sec:appendix}
\subsection{Implementation Details}

We used PyTorch for implementing the proposed method and official DeepCluster code available at \href{https://github.com/facebookresearch/swav/blob/main/main\_deepclusterv2.py}{\texttt{https://github.com/facebookresearch/swav/blob/main/main\_deepclusterv2.py}}. For easier reproducibility of our method, we provide the hyper-parameters value for our experiment on CIFAR-10 below.
\paragraph{Optimizers.} 
We used SGD for training CNN with the following hyper-parameters:
\begin{enumerate}
    \item Base Learning Rate = $6e-2$
    \item Final Learning Rate = $3e-4$
    \item Momentum = $0.9$
    \item Weight Decay = $1e-6$
    \item Epochs = $150$
    \item Batch size = $256$
    \item Warmup epochs = $5$
    \item Warmup start learning rate = $1e-6$
\end{enumerate}

\paragraph{DeepCluster hyper-parameters.}
We removed the data augmentation and muti-crop function in the original code and use following hyper-parameters:
\begin{enumerate}
    \item Number of prototypes = $60$
    \item Temperature = $0.5 $
    \item Output feature dimension = $128$
    \item Hidden layer dimension = $1024$
    \item Number of iteration with frozen prototypes = $5000$
    
\end{enumerate}

\paragraph{Self-Distillation. }
In~\autoref{eq:distill_total} of self-distillation loss, $\alpha$ and $\lambda$ were set to be $0.9$ and $1e-5$ respectively.

\paragraph{Linear Classifier. } We trained linear classifier using Adam optimizer~\cite{kingma2017adam} with learning rate~=~$1e-3$ for 200 epochs.

\end{document}